\documentclass{article}

\usepackage{arxiv}

\usepackage[utf8]{inputenc} 
\usepackage[T1]{fontenc}    
\usepackage{hyperref}       
\usepackage{url}            
\usepackage{booktabs}       
\usepackage{amsfonts}       
\usepackage{nicefrac}       
\usepackage{microtype}      
\usepackage{lipsum}		
\usepackage{graphicx}
\usepackage[numbers]{natbib}
\usepackage{doi}

\usepackage{cite}
\usepackage{amsmath,amssymb,amsfonts}
\usepackage{algorithmic}
\usepackage[ruled, longend]{algorithm2e}
\usepackage{graphicx}
\usepackage{textcomp}
\usepackage[nolist,printonlyused]{acronym}

\newacro{US}[U.S.]{United States}
\newacro{FBI}[FBI]{Federal Bureau of Investigations}
\newacro{CIA}[CIA]{Central Intelligence Agency}
\newacro{DOD}[DOD]{Department of Defense}

\begin{acronym}[SanctSound] 
\acro{HARP}{High-frequency Acoustic Recording Package}
\acro{MNLL}{mean negative log likelihood}
\acro{MASINT}{measurement and signature intelligence}
\acro{NPS}{Naval Postgraduate School}
\acro{SRWBR}{short range wide band radio}
\acro{TCP}{Transmission Control Protocol}
\acro{MARS}{Monterey accelerated research system}
\acro{MMSI}{maritime mobile service identities}
\acro{USN}{\ac{US} Navy}
\acro{NN}{neural network}
\acro{DOA}{direction of arrival}
\acrodefplural{BNN}[BNNs]{Bayesian neural networks}
\acrodefplural{GAN}[GANs]{generative adversarial networks}
\acro{AIS}{automatic identification system}
\acro{BAL}{Bayesian active learning}
\acro{UATR}{underwater acoustic target recognition}
\acro{NPS}{Naval Postgraduate School}
\acro{ML}{machine learning}
\acro{FL}{federated learning}
\acro{HFL}{hierarchical \ac{FL}}
\acro{BDL}{Bayesian deep learning}
\acro{DL}{deep learning}
\acro{CL}{continual learning}
\acro{AL}{active learning}
\acro{CNN}{convolutional neural network}
\acro{GDPR}{General Data Protection Regulation}
\acro{BFL}{Bayesian federated learning}
\acro{IID}{independent and identically distributed}
\acro{non-IID}{non-independent and identically distributed}
\acro{VI}{variational inference}
\acro{NWA}{Naive Weighted Averaging}
\acro{WS}{Weighted Sum of Normal Distributions}
\acro{LP}{Linear Pooling}
\acro{IV}{Inverse Variance Weighting}
\acro{DWC}{Distributed Weight Consolidation}
\acro{CLP}{Centered Linear Pool}
\acro{LLP}{Log-Linear Pooling}
\acro{WC}{Weighted Conflation}
\acro{BNN}{Bayesian neural network}
\acro{MC}{Monte Carlo}
\acro{KL}{Kullback-Leibler}
\acro{NLL}{negative log likelihood}
\acro{SGD}{Stochastic Gradient Descent}
\acro{ELBO}{Evidence Lower Bound}
\acro{TL}{transfer learning}
\acro{LOFAR}{low-frequency analysis and recording}
\acro{ResNet}{residual network}
\acro{mAP}{mean average precision}
\acro{AUC}{area under the curve}
\acro{VI-PANN}{variational inference pre-trained audio neural network}
\acro{PANN}{pre-trained audio neural network}
\acro{MCMC}{Markov Chain Monte Carlo}
\acro{STFT}{Short-time Fourier transform}
\acro{MFCC}{mel frequency cepstral coefficients}
\acro{CCTZ}{Chroma, Contrast, Tonnetz, and Zero-cross rate}
\acro{ResNet}{residual network}
\acro{RNN}{recurrent neural network}
\acro{NLP}{natural language processing}
\acro{LLM}{large language model}
\acro{COCO}{Common Objects in Context}
\acro{MLOps}{Machine Learning Operations}
\acro{EABO}{Expeditionary Advanced Base Operations}
\acro{EWC}{Elastic Weight Consolidation}
\acro{VCL}{Variational Continual Learning}
\acro{BALD}{Bayesian active learning by disagreement}
\acro{DARPA}{Defense Advanced Research Projects Agency}
\acro{FFT}{federated fine-tuning}
\acro{CI}{confidence interval}
\acro{OOD}{out-of-distribution}
\acro{TL}{transfer learning}
\acro{NOAA}{National Oceanic and Atmospheric Administration}
\acro{USCG}{United States Coast Guard}
\acro{SanctSound}{Sanctuary Soundscapes}
\acro{MMSI}{Maritime Mobile Service Identities}
\acro{IMO}{International Maritime Organization}
\acro{LTSA}{long-term spectral average}
\acro{UPAD}{Underwater Passive Acoustic Dataset}
\acro{SOTA}{state-of-the-art}
\acro{AI}{artificial intelligence}
\end{acronym}

\usepackage[caption=false]{subfig}
\usepackage{multirow}
\usepackage{hhline}

\title{Federated Bayesian Deep Learning: The Application of Statistical Aggregation Methods to Bayesian Models}

\date{March 22, 2024}	

\author{ \href{https://orcid.org/0000-0002-2565-6706}{John~Fischer} \\
	Department of Computer Science\\
	Naval Postgraduate School\\
	Monterey, CA 93954 \\
	\texttt{john.fischer@nps.edu} \\
	\And
	\href{https://orcid.org/0000-0003-3305-8412}{Marko~Orescanin} \\
	Department of Computer Science\\
	Naval Postgraduate School\\
	Monterey, CA 93954 \\
	\texttt{marko.orescanin@nps.edu} \\
	\And
	\href{ https://orcid.org/0009-0000-8435-7020}{Justin Loomis} \\
	Department of Computer Science\\
	Naval Postgraduate School\\
	Monterey, CA 93954 \\
	\And
    \href{https://orcid.org/0000-0002-2439-971X}{Patrick McClure} \\
	Department of Computer Science\\
	Naval Postgraduate School\\
	Monterey, CA 93954 \\
}

\renewcommand{\shorttitle}{Federated Bayesian Deep Learning}

\hypersetup{
pdftitle={Federated Bayesian Deep Learning: The Application of Statistical Aggregation Methods to Bayesian Models},
pdfauthor={John T.~Fischer, Marko~Orescanin, Justin~Loomis, Patrick~McClure},
pdfkeywords={Bayesian deep learning, federated learning, Monte Carlo dropout, uncertainty decomposition, uncertainty quantification, variational inference},
}

\begin{document}
\maketitle

\begin{abstract}
	Federated learning (FL) is an approach to training machine learning models that takes advantage of multiple distributed datasets while maintaining data privacy and reducing communication costs associated with sharing local datasets. Aggregation strategies have been developed to pool or fuse the weights and biases of distributed deterministic models; however, modern deterministic deep learning (DL) models are often poorly calibrated and lack the ability to communicate a measure of epistemic uncertainty in prediction, which is desirable for remote sensing platforms and safety-critical applications.  Conversely, Bayesian DL models are often well calibrated and capable of quantifying and communicating a measure of epistemic uncertainty along with a competitive prediction accuracy.  Unfortunately, because the weights and biases in Bayesian DL models are defined by a probability distribution, simple application of the aggregation methods associated with FL schemes for deterministic models is either impossible or results in sub-optimal performance.  In this work, we use independent and identically distributed (IID) and non-IID partitions of the CIFAR-10 dataset and a fully variational ResNet-20 architecture to analyze six different aggregation strategies for Bayesian DL models.  Additionally, we analyze the traditional federated averaging approach applied to an approximate Bayesian Monte Carlo dropout model as a lightweight alternative to more complex variational inference methods in FL. We show that aggregation strategy is a key hyperparameter in the design of a Bayesian FL system with downstream effects on accuracy, calibration, uncertainty quantification, training stability, and client compute requirements. 
\end{abstract}


\section{Introduction}

\subsection{Motivation}

Remote sensors are ubiquitous in the world today. This is due to the continual advancement in hardware technology, which supports their deployment in ever-shrinking, affordable devices.  The mass deployment of remote sensors has resulted in an explosion in the amount of data collected in support of various research groups, corporations, and government entities~\citep{dataexplosion}.  The overwhelming task of analyzing this data has resulted in a desire to automate or to augment data analysis using \ac{DL} techniques.  In many remote sensing applications, the cost of collecting, labeling, communicating, and storing local data may be acceptable; however, when deploying sensors all over the world, these costs can breach the threshold of acceptability.  In other applications, the transfer and storage of user data raises data privacy concerns and regulations, such as the \ac{GDPR}~\citep{GDPR}, and imposes significant restrictions on the handling of such data.  In order to address these issues, McMahan \textit{et al.}~\citep{McMahanFedAvg} developed a technique that they termed \ac{FL}, whereby client models train, or update, an instance of a global model using locally captured data and periodically send updated model parameters to a central server for aggregation.  By using \ac{FL}, it is possible to take advantage of numerous distributed datasets while maintaining privacy and reducing communication costs associated with training on a collection of these individual datasets. 

Since the seminal work by McMahan \textit{et al.}~\citep{McMahanFedAvg}, a significant amount of research has been done to address the numerous challenges associated with implementing an \ac{FL} framework.  This list of challenges includes, but is not limited to, efficiency of communications~\citep{communicationstrategyfedlearning, multiobjectiveevolutionaryfedlearning, robustandcommefficientfedlearning}, device heterogeneity~\citep{LiFedProx, XuDeviceHeterogeneitySampling, LaguelDeviceHeterogeneitySuperquantile}, statistical heterogeneity~\citep{ZhaoFedLearnNonIID, PFLMoreauEnvelopes, ZhangPFedBayes, PFL_ametalearningapproach}, and privacy preservation/security~\citep{PracticalSecureAggregation, DiffPrivacyFedLearning, ProtectionAgainstReconstruction, Li2018RSABS}. Often, when addressing the aforementioned problems in the \ac{FL} literature, a deterministic model architecture is assumed~\citep{McMahanFedAvg, LiFedProx, PFLMoreauEnvelopes}.  While deterministic deep learning models are capable of achieving high accuracy, they are often uncalibrated~\citep{GuoCalibrationNN} (i.e., model-predicted probabilities for each class do not match the empirical frequency for that class) and do not provide access to epistemic (model) uncertainty in prediction. The inability to access epistemic uncertainty is problematic in many remote sensing and other safety-critical applications, where it is important that the underlying model is capable of communicating a measure of epistemic uncertainty in prediction and that the uncertainty be calibrated.

Although the statistics, meteorological, and multi-sensor fusion literature are rich with probabilistic pooling or fusion approaches~\citep{MultiSensorFusionMinimumDistanceSum, CombiningProbabilityDistributions, ConflationsOfProbabilityDistributions}, there does not exist a detailed study of these probabilistic pooling/fusion strategies in the \ac{FL} setting.  Recognizing that there are numerous ways of aggregating, or fusing, probability distributions and that there is no agreed upon technique, Koliander \textit{et al.}~\citep{FusionOfPDF} provide a comprehensive review of fusion techniques and the motivation for utilizing various fusion techniques and client/site weighting schemes.  Utilizing several of the strategies from~\citep{FusionOfPDF} and additional strategies from~\citep{McMahanFedAvg} and~\citep{McClureDWC}, we analyze the effect of fusion techniques in the \ac{BFL} setting, with a focus on the quality of uncertainty.  To align with the nomenclature common in \ac{FL} literature throughout this manuscript, we use the term ``aggregate" in place of ``fuse'' or ``pool.''         

\subsection{Contributions}
In this paper, we make the following contributions: 
\begin{enumerate}
    \item We re-frame statistical aggregation strategies in the context of \ac{FL}, specifically for \ac{BDL}. 
    \item We evaluate six aggregation strategies for \ac{BDL} models in the \ac{FL} setting on \ac{IID} and non-\ac{IID} partitions of a popular computer vision dataset commonly used to benchmark performance in \ac{ML} and \ac{FL} applications. We show that aggregation strategy is a crucial hyperparameter in the design of a \ac{BFL} algorithm.
    \item We evaluate three client weighting strategies in conjunction with each of the six aggregation strategies under evaluation.
    \item We examine a popular lightweight Bayesian approximation, \ac{MC} dropout, in the \ac{FL} setting and compare its performance to more complex \ac{VI} models.
    \item We evaluate not only the prediction accuracy, but also the calibration and uncertainty (total and decomposed) of both mean-field Gaussian \ac{VI} and \ac{MC} dropout \ac{VI} models used in the \ac{FL} setting.
    \item We share our Flower Federated Learning Framework~\citep{beutel2020flower} implementation to facilitate further experimentation with aggregation methods and client weighting techniques on additional datasets.\footnote{https://github.com/meekus-fischer/BayesianFederatedLearning}
\end{enumerate}

\section{Background and Related Work}

\subsection{Federated Learning}
In \ac{FL}, client models train, or update, an instance of a global model using locally captured data and periodically send updated model parameters to a central server for aggregation~\citep{McMahanFedAvg}.  Often, this technique is utilized to take advantage of numerous local datasets while maintaining privacy.  Additionally, in remote sensing platforms and other edge devices, \ac{FL} significantly reduces the communication costs associated with training on these individual datasets.  We use the following notation throughout this manuscript: $k \in \{1,...,K\}$ represents the individual clients, $\mathcal{D}^{k} = \{(x_1,y_1), ..., (x_{|\mathcal{D}^{k}|},y_{|\mathcal{D}^{k}|})\}$ represents the local dataset of client $k$ with $|\mathcal{D}^k|$ local training examples, $\mathcal{D}$ comprises the entire training dataset split amongst $K$ clients, $\theta_{r}^{k}$ is the set of parameters for client $k$ and \ac{FL} round $r$, $\theta^{g}_{r}$ is the set of parameters for the global model at round $r$, $\omega_k$ is the aggregation weight for client $k$, B is the local minibatch size, E is the number of local epochs, and $\eta$ is the learning rate.  In general, we can describe the federated optimization problem of $K$ clients with the following:

\begin{equation}
    \underset{\theta \in \mathbb{R}^d}{\text{min}} \hspace{3pt} f(\theta^{g}), \hspace{15pt} \text{where} \hspace{20pt} f(\theta^{g}) = \sum_{k=1}^{K} \omega_{k}f_{k}(\theta^{g}).
\end{equation}
\vspace{1pt}

\noindent In the \ac{ML}/\ac{FL} context, $f_{k}(\theta)$ often corresponds to $\mathcal{L}_{k}(\mathcal{D}^{k},\theta)$, the local loss of client $k$ on $\mathcal{D}^{k}$ with model parameters $\theta$.  We place the following constraint on client aggregation weights: $\sum_{k=1}^{K}\omega_{k} = 1$.  In practice, this optimization is done, as described in Alg.~\ref{alg:Federated Learning}, over multiple optimization/communication rounds.  First, the server initializes the global model (or distributes a pre-trained baseline model) and distributes the model parameters to each of the clients.  During each optimization round $r$, clients train the global model, using a variant of \ac{SGD}, for $E$ epochs before returning the client parameters to the server.  The server then uses an aggregation function (e.g., FedAvg~\citep{McMahanFedAvg}) to combine the model weights and biases of each client into a new global model, and distributes the new global model parameters to each of the clients.  This process repeats for $R$ communication rounds, or until convergence.   

\begin{algorithm}
\caption{\label{alg:Federated Learning}Federated Learning.}

\DontPrintSemicolon

\SetKwBlock{Server}{Server Executes:}{}
\SetKwBlock{Client}{ClientUpdate($k,\theta^g$):}{}
    \SetAlgoNoLine
    \SetAlgoNoEnd
    \Server{
        initialize $\theta^{g}_{0}$\;
        
        \ForEach{round r = 0,1,...,R-1}{
            \ForEach{client $k \in \{1,...,K\}$ \textbf{in parallel}}{
                $\theta_{r+1}^{k} \leftarrow$ ClientUpdate($k,\theta^{g}_{r}$)\;
            }
            $\omega =$ CalculateClientWeights()  ... [Compute $w_{k}$ using Eq.~(\ref{eq:EqualWeight}), (\ref{eq:LocalDS}), (\ref{eq:MaxDiscWeight}), or (\ref{eq:DistFromGlobal})]\; 
            $\theta^{g}_{r+1} \leftarrow $ AggregateClients($\{\theta^{1}_{r+1},...,\theta^{K}_{r+1}\},\omega$) ... [Eq. (\ref{eq:NWA}), (\ref{eq:WS}), (\ref{eq:LP}), (\ref{eq:Conflation}), (\ref{eq:WC}), or (\ref{eq:DWC}) ]\;
        }

    }
    \Client(// Run on client $k$){
        $\mathcal{MB} \leftarrow$ (split $\mathcal{D}^{k}$ into minibatches of size \textit{B})\;
        $\theta^k \leftarrow \theta^g$\;
        \ForEach{local epoch i from 1 to E}{
            \For{minibatch b $\in \mathcal{MB}$}{
                $\theta^k \leftarrow \theta^k - \eta \nabla \ell$($\theta^k;b$)\;
            }
        }
        \textbf{return} $\theta^k$ to the server
    }
\end{algorithm}

\subsection{Bayesian Deep Learning}
To instill sufficient confidence in the performance, or capability, of \ac{ML} models for deployment on remote sensing platforms, or safety-critical applications, and to ensure the models are able to adapt to dynamic environments, it is critical that the models are capable of communicating a measure of epistemic uncertainty in prediction~\citep{Gal2016Uncertainty,Kendall_Gal_NIPS2017_2650d608}.  \ac{BDL} models apply Bayesian methods in order to provide access to the uncertainty measure (epistemic uncertainty), which deterministic \ac{ML} models are incapable of providing.  Instead of utilizing point estimates (deterministic \ac{ML}), \ac{BDL} learns a probability distribution over model parameters~\citep{blei2017variational}.  \ac{BDL} has its roots in Bayesian statistics, where inference about model parameters, $\theta$, given data, $\mathcal{D}$, involves the calculation of the posterior $p(\theta | \mathcal{D})$.  Placing a prior $p(\theta)$ on model parameters (e.g., the weights and biases of a \ac{BDL} architecture) enables the calculation of the posterior distribution of model weights and biases utilizing Bayes' rule:

\begin{equation}
    p \left(\theta|\mathcal{D} \right)= \frac{p \left(\mathcal{D}|\theta \right)p(\theta)}{p(\mathcal{D})}. 
\label{eq:bayes}
\end{equation} 
\vspace{1pt}

\noindent Ultimately, prediction amounts to calculating

\begin{equation}
    p \left(y^{*}|x^{*},\mathcal{D} \right)= \int_{\Theta}^{} p \left(y^{*}|x^{*},\theta \right) p \left(\theta|\mathcal{D}\right)d\theta,
\label{eq:inference}
\end{equation} 
\vspace{1pt}

\noindent where $x^{*}$ is the input feature vector, $y^{*}$ is a corresponding prediction, $\Theta$ represents the parameter space of the model, and $p\left(y^{*}|x^{*},\mathcal{D} \right)$ is the probability of that prediction given a trained model.  Due to the fact that $p(\mathcal{D})$ in Eq.~(\ref{eq:bayes}) is not directly accessible, it must be calculated using known quantities.

\begin{equation}
    p(\mathcal{D}) = \int_{\Theta}^{}p(\mathcal{D},\theta)d\theta.
\label{eq:joint}
\end{equation}
\vspace{1pt}

High-dimensional integrals, such as the one in Eq.~(\ref{eq:joint}), often have no closed form and are computationally intractable~\citep{probDeepLearn}.  In practice, approximate inference is used.  There are many methods for approximate inference; however, two of the most common approaches are \ac{MCMC} sampling and \ac{VI}~\citep{bishop2006pattern}.  Although \ac{MCMC} is often described as the ``gold standard'' for approximate inference, \ac{VI} is commonly favored in \ac{BDL}, due to the increased speed and the ability to scale with large data and large models~\citep{JMLR:v14:hoffman13a}. There are a number of approaches to implementing VI; two of the most common of implementations---Flipout~\citep{wen2018flipout} and Local Reparameterization Trick~\citep{kingma_variational_2015}---use Gaussian distributions and effectively double the number of model parameters. Another common approach, \ac{MC} dropout, was shown to be an effective Bayesian approximation that does not double the number of model parameters~\citep{gal2016dropout}.  In many situations, each of these methods has been empirically shown to lead to useful approximation of $p(\theta |\mathcal{D})$, providing access to improved uncertainty estimation when compared to the uncertainty provided by deterministic models.  In this work, we utilize Flipout \ac{VI} models and \ac{MC} dropout models as our \ac{BDL} models.  

The essence of \ac{VI} is that optimization is used to approximate the true posterior $p(\theta|\mathcal{D})$ by identifying the closest distribution $q_{\phi}(\theta)$, parameterized by $\phi$, to the posterior among a family of pre-determined (variational) distributions (commonly a Gaussian).  Often, the ``closeness'' is determined using \ac{KL} divergence, a common information-theoretic measure of similarity between two distributions.

\begin{equation}
    KL[ q_{\phi}(\theta)\mid\mid p(\theta|\mathcal{D})] =\int_{\Theta}^{}q_{\phi}(\theta)log \frac{q_{\phi}(\theta)}{p(\theta|\mathcal{D})}d\theta.
\label{eq:KL}
\end{equation}
\vspace{1pt}

\noindent The optimization objective for \ac{VI} then becomes

\begin{equation}
    \mathcal{L}_q = \text{KL}[q_{\phi}(\theta)\mid\mid p(\theta)] - \mathbb{E}_{q} \left[\log p\left(\mathcal{D}|\theta\right) \right], 
\label{eq:ELBO}
\end{equation}
\vspace{1pt}

\noindent where $\mathbb{E}_q$ represents the expected value under the probability distribution $q_{\phi}(\theta)$.  In this form, also known as the negative \ac{ELBO}, it becomes clear that the optimization objective is the sum of a data-dependent portion (likelihood cost), and a prior dependent portion (complexity cost)~\citep{pmlr-v37-blundell15}. 

The quality of the approximation in \ac{VI} is highly dependent on the choice of the variational distribution family.  As mentioned in the previous paragraph, we consider two distinct approaches, a mean-field Gaussian distribution with a Flipout \ac{MC} estimator of \ac{KL}-divergence~\citep{wen2018flipout} and \ac{MC} dropout, which uses a Bernoulli distribution over the model weights and biases~\citep{gal2016dropout}.  In the first, the use of Gaussian distributions, parameterized by the mean and variance of the distribution, doubles the number of parameters in the model architecture, when compared to its deterministic counterpart.  The use of the Gaussian distribution results in a more challenging optimization problem relative to \ac{MC} dropout.  In the \ac{MC} dropout approach, there is no additional increase in the number of model parameters, and minimal changes are required to the model architecture~\citep{gal2017bald, Gal2016Uncertainty, gal2016dropout}.  It is these simplifications that motivate our analysis of the \ac{MC} dropout model in the \ac{FL} framework as a lightweight Bayesian approximation.

\subsection{Uncertainty Quantification and Decomposition}
In \ac{BDL} models used for classification, predictive probability $p(y = c \mid x)$ is approximated by using \ac{MC} integration with $M$ samples~\citep{filos2019systematic}.  The mean probability per class $\bar{p}_{c}$ is calculated using the prediction probabilities, $\hat{p}_{c_{m}}$, where $\hat{p}_{c_{m}}= p(y = c \mid x,\theta^m)$ and $\theta^m$ is sampled from an approximation of $p(\theta|\mathcal{D})$.

\begin{equation}
    \bar{p}_{c} = \frac{1}{M}\sum_{m=1}^{M} \hat{p}_{c_{m}}.
\label{eq:mean_prob}
\end{equation}
\vspace{1pt}

The predicted class for each sample is chosen by selecting the class that yields the highest mean probability.  In addition to a predicted class, the $M$ \ac{MC} samples allow for the calculation of predictive entropy and variance.  Predictive entropy has several interpretations; however, in information theory, entropy is interpreted as the amount of information contained in a predictive distribution \citep{probDeepLearn}.  Normalized predictive entropy ($\tilde{H}$), for the multi-class classification problem, is given by

\begin{equation}
    \tilde{H}_{p}(y \mid \textbf{x}) = - \sum_{c \in C}  p(y = c \mid x) \frac{ \log p(y = c \mid x)}{\log C},
\label{eq:pred_entropy}
\end{equation}
\vspace{1pt}

\noindent where $C$ represents all possible classes~\citep{park_using_2015}.  Although entropy is commonly used as a measure of aleatoric uncertainty in deterministic and Bayesian deep learning, Depeweg \textit{et al.} \citep{pmlr-v80-depeweg18a} interpret entropy as a measure of predictive uncertainty, which could be further broken down into its aleatoric and epistemic uncertainty components. Aleatoric uncertainty is known as data uncertainty and cannot be reduced with an increase in training data.  Epistemic uncertainty is known as model uncertainty and can be reduced by increasing the amount of training data~\citep{probDeepLearn}. Epistemic uncertainty is not accessible using deterministic models.

For \ac{BDL} models, variance provides another measure of total uncertainty, which can be broken down into epistemic and aleatoric uncertainty.  Kendall \textit{et al.}~\citep{Kendall_Gal_NIPS2017_2650d608} and Kwon \textit{et al.}~\citep{kwon2020uncertainty} separately propose methods for estimating aleatoric and epistemic uncertainty with Kwon \textit{et al.} modifying the approach in~\citep{Kendall_Gal_NIPS2017_2650d608} to give the following estimator:

\begin{equation}
   \underbrace{\frac{1}{M}\sum_{m=1}^{M} \textnormal{diag}(\hat{p}_{m})-\hat{p}_{m}^{\otimes2}}_{aleatoric} + \underbrace{\frac{1}{M}\sum_{m=1}^{M} (\hat{p}_{m}-\bar{p}_{c})^{\otimes2}}_{epistemic},
\label{eq:aleatoricEpistemic}
\end{equation}
\vspace{1pt}

\noindent where $M$ is the number of \ac{MC} samples, $\hat{p}_{m}$ is a $c$-dimensional vector containing the predictive probabilities of each class ($\hat{p}_{{c}_{m}}$), $\textnormal{diag}(\hat{p}_m)$ is a diagonal matrix, ${\hat{p}_m}^{\otimes 2} = \hat{p}_m{\hat{p}_m}^T$, and $({\hat{p}_m} - \overline{p}_{c})^{\otimes 2} = ({\hat{p}_m} - \overline{p}_{c})({\hat{p}_m} - \overline{p}_{c})^T$.

\subsection{Federated Bayesian Learning}

In scenarios involving big data with many distributed remote sensing platforms, the need to adapt to dynamic environments, and the need to benefit from the ability to communicate a measure of uncertainty, the application of \ac{BDL} to \ac{FL} is a natural one. Several works have applied Bayesian techniques to the \ac{FL} problem.  In~\citep{LiuFederatedLaplace}, the authors apply the Laplace approximation at both the client and server to reduce aggregation errors and regularize, or guide, local model training.  Notably, in this work, the authors choose to aggregate client models using a mixture of Gaussians.  FedBE~\citep{ChenChaoFedBE} constructs a global model by fitting a distribution (Gaussian or Dirichlet) to the client models and sampling from this distribution to arrive at a higher-quality global model.  Similarly, FedAG~\citep{ProbabilisticPredictionsFederatedLearning} applies uncertainty in the aggregation step by using the client parameters to fit a Gaussian distribution at the server.  In~\citep{BhattBFLviaPDD}, the authors present a \ac{BFL} framework where each client performs approximate Bayesian inference, using MCMC sampling, and distills the clients' posterior distribution into a single deep neural network via knowledge distillation techniques for aggregation at the server.  pFedBayes~\citep{ZhangPFedBayes} is a personalized \ac{FL} model that uses \ac{VI} to learn a personalized local model at each client while learning a shared global model.  

With the exception of pFedBayes~\citep{ZhangPFedBayes}, none of these works employ \ac{VI}.  While pFedBayes~\citep{ZhangPFedBayes} utilizes \ac{VI}, the global model uses a simple averaging method for client parameter aggregation, and no additional methods are discussed, or analyzed.  Additionally, in pFedBayes~\citep{ZhangPFedBayes}, the authors utilize a model with a single variational layer (i.e., with the exception of the output layer, all layers are deterministic).  This work is the first time, to our knowledge, that anyone has utilized a fully variational model (i.e., all model layers are variational) to analyze the effects of the choice of aggregation function and client weighting scheme on \ac{BDL} models in \ac{FL}.

The use of \ac{BDL} in \ac{FL} introduces several unique problems.  First, the model weights and biases in a \ac{VI} model are no longer a point estimate.  Each weight and bias is defined by the parameters of a distribution.  In the case of a Gaussian, this is the mean and variance of the distribution.  In the following sub-sections, we present multiple aggregation strategies for univariate Gaussian distributions and their underlying theory.  Next, we discuss the selection of client weights in the aggregation process.  Often, the size of the local training dataset is utilized to weight each client. However, this is not the only method, and we present several other weighting schemes for analysis.  Finally, in mean-field Gaussian \ac{VI} we often select priors with a fixed mean of zero and a fixed variance, $p(\theta) = N(0,\sigma_{prior}^2I)$, for the regularization term in the optimization problem. However, in this work, we pose the question of whether or not to update the prior after each round of optimization.  We analyze the effect of updating the prior each round vs. maintaining a fixed prior throughout the training process.

\subsection{Aggregation Strategies}
Although there are a significant number probabilistic aggregation approaches that have been presented in the literature~\citep{MultiSensorFusionMinimumDistanceSum, CombiningProbabilityDistributions, ConflationsOfProbabilityDistributions}, we focus our study on six select strategies from~\citep{McMahanFedAvg,ConflationsOfProbabilityDistributions,FusionOfPDF,HillCombiningDatasets,InverseVarianceWeighting} due to their suitability for implementation in the \ac{FL} construct.  In the following sub-sections, we discuss each of these approaches using the univariate Gaussian distribution for each parameter ($\theta_i$), parameterized by its mean ($\mu_i$) and variance ($\sigma_{i}^2$), as the model distribution. We use univariate Gaussians because of our use of mean-field Gaussian \ac{VI}, where we assume each parameter is independent. The six approaches are \ac{NWA}~\citep{McMahanFedAvg}, \ac{WS}, \ac{LP}~\citep{FusionOfPDF}, Conflation/\ac{WC}~\citep{ConflationsOfProbabilityDistributions,HillCombiningDatasets}, and \ac{DWC}~\citep{McClureDWC}.  We use the following notation in our discussion: $K$ is the set of client models, $q_k$ is the distribution for client $k$, $q_g$ is the global (server) distribution, and $\omega_k$ is the weight of client $k$ with $\sum_{k=1}^{K} \omega_{k} = 1$.

\subsubsection{Naive Weighted Averaging}
This aggregation method is similar to the FedAvg algorithm~\citep{McMahanFedAvg} and is simplest of the aggregation techniques.  In the case where $\omega_k = \frac{|\mathcal{D}_k|}{|\mathcal{D}|}$, \ac{NWA} is equivalent to FedAvg.  Although this method was shown to be effective for aggregating deterministic model weights and biases in~\citep{McMahanFedAvg}, independently aggregating the mean and variance parameters in this manner ignores the statistical properties of the Gaussian Distribution.  This method is familiar to the statistical literature, however, and is described as \ac{CLP} in~\citep{KnuppelForecastUncertainty}, and used extensively in~\citep{KnuppelForecastUncertainty,clements2018densityforecasts,zarnowitzconsensus}.  Clements defends the use of this variance calculation in~\citep{clements2018densityforecasts} noting that the variance calculated using \ac{LP} is often inflated by the ``disagreement'' term (described in Section II.G.3), and the \ac{NWA} method is equal to the expected variance of a randomly selected client.       

\begin{equation}\label{eq:NWA}
    \mu_{q_{g}} = \sum_{k=1}^{K} \omega_{k}\mu_{q_k}, \hspace{10pt}
    \sigma_{q_{g}}^2 = \sum_{k=1}^{K} \omega_{k}\sigma_{q_k}^2.
\end{equation}    
\vspace{1pt}

Ultimately, this method involves taking the weighted average of each of the clients' individual model parameters to form the global model without regard for what each of the parameters represents.

\subsubsection{Weighted Sum of Normal Distributions}
If the mean and variance of each model parameter are considered as part of a univariate Gaussian distribution, instead of being considered individually as in \ac{NWA}, it is possible to take the weighted average of each of the client Gaussian distributions.  This strategy assumes the weights of each client are mutually independent. 

\begin{equation}\label{eq:WS}
        \mu_{q_{g}} = \sum_{k=1}^{K} \omega_{k}\mu_{q_{k}}, \hspace{10pt}
        \sigma_{q_{g}}^2 = \sum_{k=1}^{K} \omega_{k}^2\sigma_{q_{k}}^{2}.
\end{equation}
\vspace{1pt}

The resultant Gaussian distribution in \ac{WS} has the same mean as in \ac{NWA}, but the variance parameter differs.  Because the $\omega$ parameter falls in the range of $[0,1]$, and we square each client's weight parameter $\omega_k$ in \ac{WS}, the variance parameter in \ac{WS} will always be less than or equal to the variance parameter in \ac{NWA}.

\subsubsection{Linear Pooling}
The \ac{LP} technique is studied extensively in the forecasting literature~\citep{theopinionpool,KnuppelForecastUncertainty,zarnowitzconsensus,clements2014forecastuncertainty,lahiriForecastUncertainty,Genest1990AllocatingTW} and is connected to model averaging and multiple model adaptive estimation (MMAE)~\citep{MMAE}, as described by Koliander \textit{et al.} in~\citep{FusionOfPDF}.  To derive the \ac{LP} operator for univariate Gaussian random variables, we define each client distribution as $q_{k}{(\theta)} = \mathcal{N}(\theta;\mu_{q_{k}},\sigma_{q_{k}})$ where $\mu_{q_{k}} = \mathbb{E}_{q_{k}}[\theta]$ and $\sigma_{q_{k}}^2 = \mathbb{E}_{q_{k}}[(\theta - \mu_{q_{k}})^2]$.  \ac{LP} is then defined as the linear combination of each of the client distributions:

\begin{equation}\label{eq:LP}
    q_{g}{(\theta)} = \sum_{k=1}^{k} \omega_{k}\mathcal{N}(\theta;\mu_{q_{k}},\sigma_{q_{k}}).
\end{equation}
\vspace{1pt}

The global mean is then defined by $\mu_{q_{g}} = \mathbb{E}_{q_{g}}[\theta]$, which is just the weighted average of the client means ($\mu_{q_{k}}$):

\begin{equation}\label{eq:LPmean}
    \mu_{q_{g}} = \sum_{k=1}^{K} \omega_{k}\mu_{q_{k}}.
\end{equation}
\vspace{1pt}

The global variance can similarly be defined as $\sigma_{q_{g}}^2 = \mathbb{E}_{q_{g}}[(\theta - \mu_{q_{g}})^2]$, which is broken down by Lahiri \textit{et al.}~\citep{lahiri1988} into two components.  The first component is the weighted average variance of each client $\sigma_{q_{k}}^2$.  The second component is the ``disagreement'' component of the global variance and captures the discrepancy between clients.

\begin{equation}\label{eq:LPvar}
    \begin{split}
        \sigma_{q_{g}}^2 = \underbrace{\sum_{k=1}^{K} \omega_{k} \sigma_{q_{k}}^2}_{\text{avg client var}} + \underbrace{\sum_{k=1}^{K} \omega_{k}(\mu_{q_{k}} - \mu_{q_{g}})^2}_{\text{disagreement term}} \\
         = \sum_{k=1}^{K} \omega_{k}(\sigma_{k}^2 + (\mu_{k} - \mu_{q_{g}})^2).
    \end{split}
\end{equation}
\vspace{1pt}

In~\citep{KnuppelForecastUncertainty}, the authors show that \ac{LP} imposes an upward bias on the global variance resulting in under-confident predictions.  We briefly mentioned one of the ways in which researchers have dealt with this upward bias in the \ac{NWA} subsection. \ac{CLP}, which is equivalent to \ac{NWA}, is a variant of \ac{LP} which ignores the disagreement term in the \ac{LP} variance calculation and reduces this upward bias.  

\subsubsection{Conflation and Weighted Conflation}
The Conflation and \ac{WC} methods~\citep{ConflationsOfProbabilityDistributions,HillCombiningDatasets} are the normalized product of client distributions and are closely related to \ac{LLP}~\citep{FusionOfPDF}.  \ac{LLP} is a weighted geometric average of client distributions and is so named because it is a linear function of client distributions in the log-domain~\citep{FusionOfPDF}.  When combining Gaussian distributions, \ac{LLP} differs from \ac{WC} only in the numerator of the global variance term, with \ac{LLP} replacing $\omega_{max}$ with a value of 1. As shown in~\citep{ConflationsOfProbabilityDistributions, HillCombiningDatasets}, Conflation/\ac{WC} has several desirable properties.  These properties include: 
\begin{itemize}
    \item Conflation/\ac{WC} is commutative, associative, and iterative.
    \item The resulting distribution of the Conflation/\ac{WC} of Gaussian distributions is a Gaussian.
    \item The result of the Conflation/\ac{WC} of Gaussian distributions matches the result of the weighted-least-squares method. 
    \item Conflation/\ac{WC} minimizes the loss of Shannon information as a result of the consolidation of multiple Gaussian distributions into a single Gaussian distribution.
    \item Conflation/\ac{WC} is the best linear unbiased estimate.
    \item Conflation/\ac{WC} yields a maximum likelihood estimator.
\end{itemize}
We provide the formulas for both Conflation, Eq.~(\ref{eq:Conflation}), and \ac{WC}, Eq.~(\ref{eq:WC}), below.

\begin{align}\label{eq:Conflation}
        \mu_{q_{g}} = \frac{\sum_{k=1}^{K} \frac{\mu_{q_k}}{\sigma_{q_k}^2}}{\sum_{k=1}^{K} \frac{1}{\sigma_{q_k}^2}}, \hspace{10pt}
        \sigma_{q_{g}}^2 = \frac{1}{\sum_{k=1}^{K} \frac{1}{\sigma_{q_k}^2}}.
\end{align}
\vspace{1pt}

Readers familiar with the \ac{IV} technique~\citep{InverseVarianceWeighting} may notice that the Conflation formula, Eq.~(\ref{eq:Conflation}), is identical to the formula for \ac{IV}.  \ac{IV} is a well-known method for combining random variables in such a way as to minimize the variance of the resulting distribution.  

\begin{align}\label{eq:WC}
        \mu_{q_{g}} = \frac{\sum_{k=1}^{K} \frac{\omega_{k}\mu_{q_k}}{\sigma_{q_k}^2}}{\sum_{k=1}^{K} \frac{\omega_k}{\sigma_{q_k}^2}}, \hspace{10pt}
        \sigma_{q_{g}}^2 = \frac{\omega_{max}}{\sum_{k=1}^{K} \frac{\omega_k}{\sigma_{q_k}^2}}.
\end{align}
\vspace{1pt}

Notably, \ac{WC} produces a Gaussian distribution with a mean value which is proportional to the client weights, inversely proportional to client variances, and has a variance which is never greater than the variance of the client with the greatest weight.

\subsubsection{Distributed Weight Consolidation} 
\ac{DWC} was introduced by McClure \textit{et al.}~\citep{McClureDWC} as a continual learning technique to combine or consolidate the model weights and biases of multiple \acp{BNN} trained on independent datasets.  This method builds on the concept of Variational Continual Learning~\citep{variationalcontinuallearning} and Bayesian Incremental Learning~\citep{bayesianincrementallearning}. However, both of these methods require that the underlying models be trained sequentially.  In \ac{DWC}, multiple disparate models (clients) can be trained independently, and their weights and biases can be consolidated into a single network (global model), similar to the \ac{FL} setting. \ac{DWC} is related to the Supra-Bayesian framework described in~\citep{FusionOfPDF}, which involves the selection of a prior distribution, $p(\theta)$, and the update of the prior using a Bayesian update rule. In \ac{DWC}, we are using the global model, $q_{\circ}(\theta)$, as the prior, $p(\theta)$, and applying the same Bayesian update rule. In Eq.~(\ref{eq:DWC}), we provide the formulas for the global mean and variance, the derivation of which can be found in~\citep{McClureDWC}.

\begin{equation}\label{eq:DWC} 
        \mu_{q_{g}} = \frac{\sum_{k=1}^{K} \frac{\mu_{q_{k}}}{\sigma_{q_{k}}^2} - \sum_{k=1}^{K-1} \frac{\mu_{q_{\circ}}}{\sigma_{q_{\circ}}^{2}}}
                            {\sum_{k=1}^{K} \frac{1}{\sigma_{q_{k}}^2} - \sum_{k=1}^{K-1} \frac{1}{\sigma_{q_{\circ}}^{2}}}, \hspace{10pt}
        \sigma_{q_{g}}^2 = \frac{1}{\sum_{k=1}^{K} \frac{1}{\sigma_{q_{k}}^2} - \sum_{k=1}^{K-1}\frac{1}{\sigma_{q_{\circ}}^{2}}}. 
\end{equation}
\vspace{1pt}

where $\mu_{q_{\circ}}$ and $\sigma_{q_{\circ}}^{2}$ correspond to the global mean and variance from the previous round, respectively.   It should be noted that \ac{DWC} was designed to start with a base model trained on some initial dataset, and not a random initialization.  Additionally, in the \ac{DWC} setting, each ``client'' trains asynchronously on their local dataset before aggregating client parameters. This is the first time that this method has been applied to and analyzed in the \ac{BFL} setting.

\subsection{Determining Weight Parameters}
In the seminal work on \ac{FL}, McMahan \textit{et al.}~\citep{McMahanFedAvg} weighted client updates based on the size of each client's local dataset relative to the datasets of all other clients.  Although this method is common in \ac{FL}, there have been several works~\citep{precisionweightedFL, FusionOfPDF} which explore various techniques for weighting clients to achieve varying objectives (robustness against outliers, reducing communications, training efficiency, etc.). We treat the client weighting scheme as a hyperparameter and analyze the results (with a focus on uncertainty) of using four different techniques for selecting client weights.  Notably, the implementation of several of these methods require that the underlying models be Bayesian.  As such, we only apply each weighting technique to models for which it is applicable.

\subsubsection{Equal Weight}
The first, and simplest, method is to give equal weight to each client in the update process using Eq.~(\ref{eq:EqualWeight}).  In addition to its simplicity, it does not require any a priori knowledge of the client datasets.  As a result, it is completely independent of the amount of training data, or the quality of training data, at each client site, and is not resistant to outliers or extreme updates.  This method is applicable to both deterministic and probabilistic models.

\begin{equation}\label{eq:EqualWeight}
        \omega_{k} = \frac{1}{K}.
\end{equation}
\vspace{1pt}

\subsubsection{Local Train Dataset Size}
This method was introduced in~\citep{McMahanFedAvg}, and is widely used in the \ac{FL} literature.  In order to apply this weighting scheme, the server either needs to have a priori knowledge of the size of the client datasets, or the client needs to communicate the size of their training dataset with the model update.  When the data is distributed evenly amongst clients, this method, Eq.~(\ref{eq:LocalDS}), is equivalent to Equal Weight.  This method is applicable to both deterministic and probabilistic models.  Since each of our dataset partitions results in clients having an equal amount of data, this method is equivalent to Equal Weight.

\begin{equation}\label{eq:LocalDS}
        \omega_{k} = \frac{|\mathcal{D}_{k}|}{|\mathcal{D}|}.
\end{equation}
\vspace{1pt}

\subsubsection{Maximum Discrepancy Weighting}
Koliander \textit{et al.} described maximum discrepancy weighting in~\citep{FusionOfPDF}.  This technique involves treating the client weights as a measure of distance.  In Eq.~(\ref{eq:MaxDiscWeight}), we use Kullback Leibler Divergence. However, other measures of distance or divergence could be used.  By computing client weights in this manner, the server places higher importance on model weights and biases that are not extreme (relative to the other clients).  The gamma value ($\gamma_k$) and omega value ($\omega_k$) in Eq.~(\ref{eq:MaxDiscWeight}) are the un-normalized and normalized weights of each client, respectively.  This method is only applicable to probabilistic models. However, a similar technique could be used in deterministic models using the L2 norm, or another measure of distance.  We only apply this technique to the probabilistic (\ac{VI}) models.

\begin{equation}\label{eq:MaxDiscWeight}
        \gamma_{k} = \underset{j \in \{1,...,K\}}{\mathrm{max}}\frac{1}{\text{KL}[\, q_{k}\mid\mid q_{j} ]\,},  
        \hspace{10pt}
        \omega_{k} = \frac{\gamma_{k}}{\sum_{j=1}^{K} \gamma_{j}}.
\end{equation}
\vspace{1pt}

\subsubsection{Distance to a Fixed Point}
This method is similar to maximum discrepancy weighting in that it also treats client weights as a measure of distance.  Instead of calculating the divergences between the distributions of the model weights and biases for individual clients, we use the divergence between the distributions of the model weights and biases for individual clients and the global model from the previous round.  In effect, this places a greater importance on clients that do not diverge significantly from the current global model.  Similar to maximum discrepancy weighting, the gamma value ($\gamma_k$) and omega value ($\omega_k$) in Eq.~(\ref{eq:DistFromGlobal}) are the un-normalized and normalized weights of each client, respectively.  Again, this method is only applicable to probabilistic models; however, a similar technique could be used in deterministic models using the L2 norm, or another measure of distance.  We only apply this technique to the probabilistic (\ac{VI}) models.

\begin{equation}\label{eq:DistFromGlobal}
        \gamma_{k} = \frac{1}{\text{KL}[\, q_{g}\mid\mid q_{k} ]\,}, 
        \hspace{10pt}
        \omega_{k} = \frac{\gamma_{k}}{\sum_{j=1}^{K} \gamma_{j}}.
\end{equation}
\vspace{1pt}

\section{Experiments}

\subsection{Datasets}
We utilize the CIFAR-10~\citep{krizhevsky2009cifar} image classification dataset to benchmark our models and model parameter aggregation rules. The CIFAR-10 dataset consists of 50,000 training images and 10,000 test images of size 32x32x3 pixels, which belong to one of 10 classes. Similar to~\citep{McMahanFedAvg,ZhaoFedLearnNonIID,LiDataSilos}, we adopt both an \ac{IID} and non-\ac{IID} partitioning strategy.  In both strategies, we partition the training dataset amongst 10 clients.  For the IID strategy, we assign each client a uniform distribution over the 10 classes.  For the non-\ac{IID} strategy, we follow the strategy introduced by Li \textit{et al.} in~\citep{LiDataSilos} and adopt both a quantity-based label imbalance and a distribution-based label imbalance.  In the quantity-based label imbalance scenario, we first sort the data by class and divide the data into 20 partitions of size 2,500, and randomly assign each client two partitions from different classes (i.e., each client training dataset contains examples from  two classes with 2,500 examples of each class).  In the distribution-based label imbalance scenario, each client is allocated a proportion of the samples of each label according to the Dirichlet distribution, which is parameterized by $\alpha$~\citep{DirichletFedLearn}. In our experiment, we use $\alpha$ parameters of 0.5 and 5.0 to simulate different levels of class imbalance across client training datasets.

\subsection{Architecture}
\begin{figure*}[ht!]
    \centering
    \includegraphics[width=0.98\linewidth]{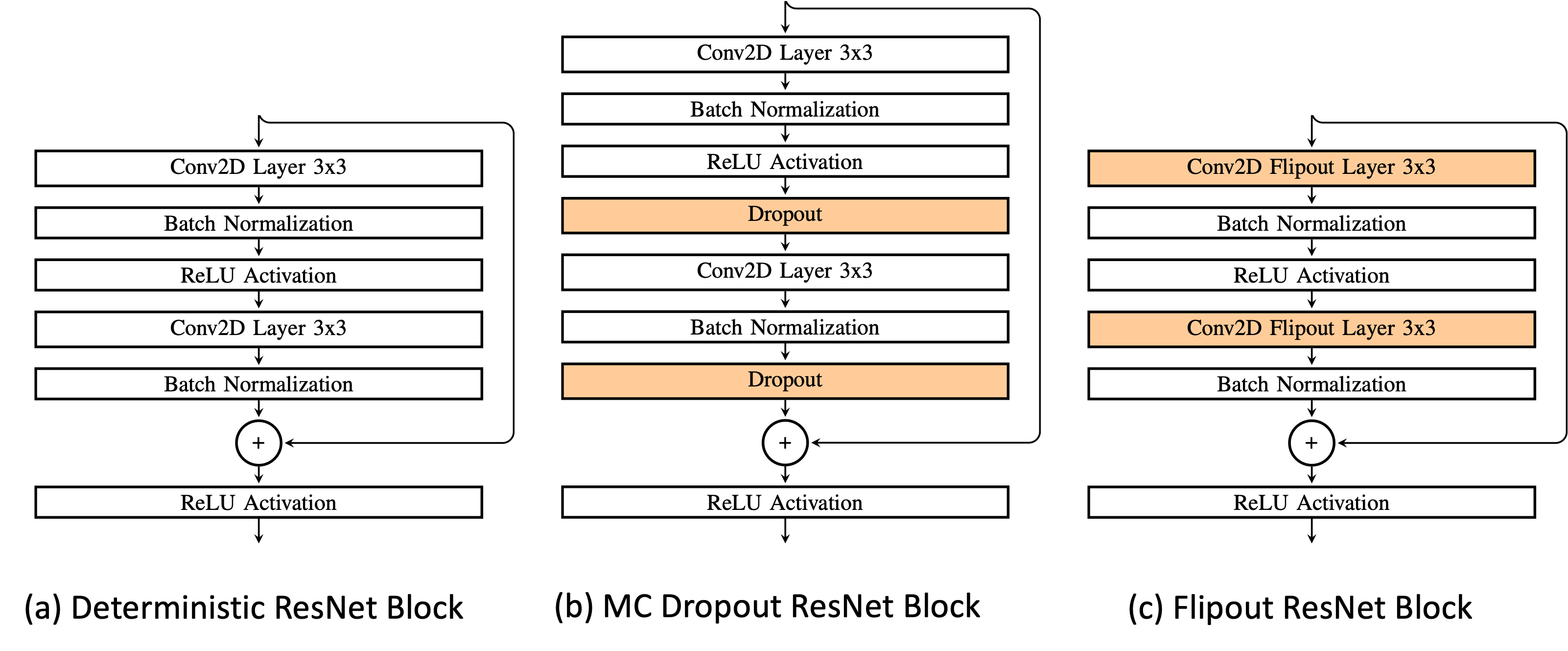}
    
    \caption{\label{fig:ResNetModelArchitecture}{ResNet Model Block Structure.  (a) depicts the structure of a ResNet block in the deterministic configuration.  (b) depicts the structure of the same ResNet block in the MC dropout configuration with the addition of a dropout layer after each convolutional and linear layer.  These dropout layers are enabled during the inference phase. (c) depicts the ResNet block in the Flipout configuration.  Each convolutional and linear layer is replaced by the Tensorflow Probability implementation of the corresponding Flipout Layer.  Changes from the deterministic configuration are highlighted in orange.}}
    
\end{figure*}

We utilize the Flower Federated Learning Framework~\citep{beutel2020flower} to implement the \ac{FL} setup. In order to conduct a fair comparison of performance across model configurations, we utilize the ResNet-20~\citep{he2016deep} architecture in a deterministic, \ac{MC} dropout~\citep{gal2016dropout}, and Flipout~\citep{wen2018flipout} (VI) configuration, as shown in Fig.~\ref{fig:ResNetModelArchitecture}. Notably, when trained on a centralized training dataset, all variants of the ResNet-20 model used in this paper achieved comparable performance to the benchmark performance reported by He \textit{et al.}~\citep{he2016deep} on the CIFAR-10 dataset, see Table~\ref{tab:centralizedresults}.  The models were coded using Tensorflow v2.9 and Tensorflow Probability v0.16~\citep{tensorflow2015-whitepaper} and all experiments were performed on NVIDIA RTX 8000 graphics processing units.

\begin{table}[ht!]
    \small
    \begin{center}
    \caption{ResNet-20 Centralized Training Results}
    \label{tab:centralizedresults}
        \begin{tabular}{|c | c |}
            \hline
        
            \textbf{Model} & \textbf{Accuracy} \\
        
            \hline
        
            Deterministic &  90.14 \\

            \hline
        
            \ac{MC} Dropout & 90.91 \\

            \hline
        
            Flipout &  90.42 \\

            \hline
        
        \end{tabular}
  \end{center}
\end{table}

\subsection{General Experiment Setup and Evaluation Metrics}
For all experiments, we utilize $K = 10$ clients, $B = 32$ local minibatch size, and $\eta = .001$ initial local learning rate with an exponential schedule that reduces the learning rate by a factor of .04 every round after a silent period of 50 aggregation rounds.  Additionally, we use a dropout rate of 0.2 for \ac{MC} dropout models and our Flipout models use $p(\theta) = N(0,100 I)$.  

We present the results of each experiment using accuracy and \ac{NLL}.  \ac{NLL} is equivalent to the KL Divergence between a model's predicted distribution and a one-hot encoded true distribution~\citep{probDeepLearn}, with a lower \ac{NLL} corresponding to a better representation.  In terms of calibration, Filos \textit{et al.}~\citep{filos2019systematic} demonstrated that if a model's performance increases as the number of discarded high-uncertainty predictions increases, the model is well-calibrated. As a result, we utilize accuracy vs. data retained curves to assess the calibration of models based on normalized predictive entropy, aleatoric uncertainty, and epistemic uncertainty.

\subsection{Evaluation of Aggregation Methods - 1 Local Epoch}
In this experiment, we evaluate each of the proposed aggregation strategies with $E = 1$ local epochs.  Each simulation is run three times for 400 rounds, and the results, shown in Table~\ref{tab:resultsLocalEpoch1}, are averaged across experiments.  

In terms of accuracy and \ac{NLL}, no aggregation strategy outperforms other aggregation strategies on all data distributions.  Notably, \ac{NWA} and \ac{LP} do not perform as well as \ac{WS}, \ac{WC}, and conflation, which all perform similarly to the deterministic model baseline (\ac{NWA}).  \ac{LP} performs especially poorly on all data distributions.

\begin{figure*}[!ht]
    \centering
        \subfloat{
            \includegraphics[width=.48\textwidth]{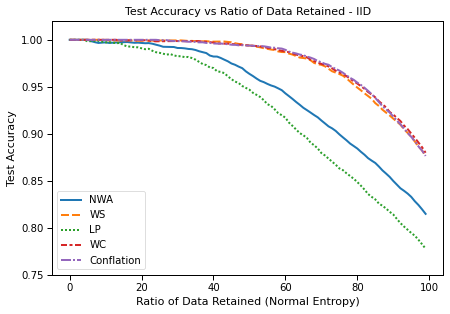}
        }
        \subfloat{
            \includegraphics[width=.48\textwidth]{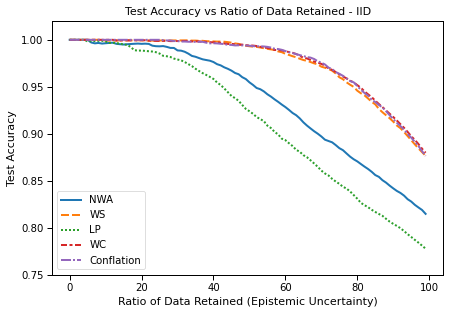}
        }\hfill %
        \subfloat{
            \includegraphics[width=.48\textwidth]{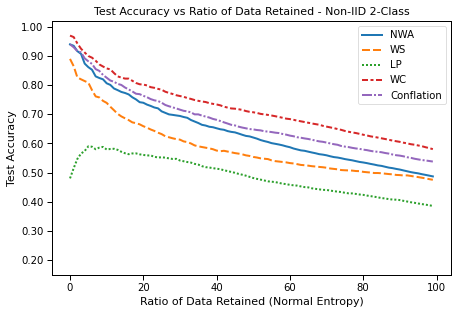}
        }
        \subfloat{
            \includegraphics[width=.48\textwidth]{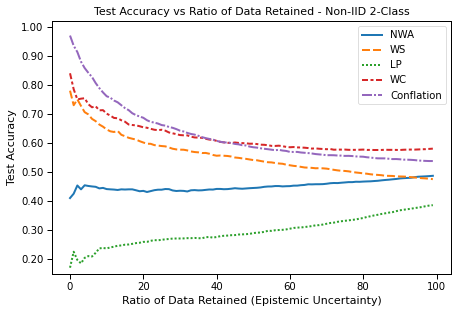}
        }\hfill%

    \caption{Uncertainty calibration plots for $E=1$.  \ac{VI} aggregation method comparison plot of test set accuracy vs. ratio of data retained for \ac{IID} and non-IID 2-class data distributions. Non-IID Dirichlet distribution plots show trends and relationships that mirror IID distribution results. Normalized entropy and epistemic uncertainty plots are shown, aleatoric uncertainty plots show trends and relationships that mirror normalized entropy.  }  
    \label{fig:E1_calibrationcomparison}
\end{figure*}

Uncertainty calibration plots are depicted in Fig.~\ref{fig:E1_calibrationcomparison}.  When looking at the bulk uncertainty metric (normalized entropy), in the \ac{IID} and non-IID Dirichlet (0.5 and 5.0) cases, all of the global models display a calibrated uncertainty.  Even though the global models are calibrated, the models trained using \ac{WS}, \ac{WC}, and conflation appear to be better calibrated than the models trained using \ac{LP} and \ac{NWA} (i.e., test accuracy increases more rapidly as the most uncertain samples are discarded from the test data).  By breaking down the bulk uncertainty into its epistemic and aleatoric components, the performance difference is amplified.  Specifically, when filtering based on epistemic uncertainty, models trained using \ac{WS}, \ac{WC}, and conflation perform much better than those trained using \ac{LP} and \ac{NWA}.  In the non-IID 2-Class case, the relationships between aggregation strategies is similar. However, not all the global models have a calibrated epistemic uncertainty component.  

\begin{figure*}[!ht]
    \centering
        \includegraphics[width=\textwidth]{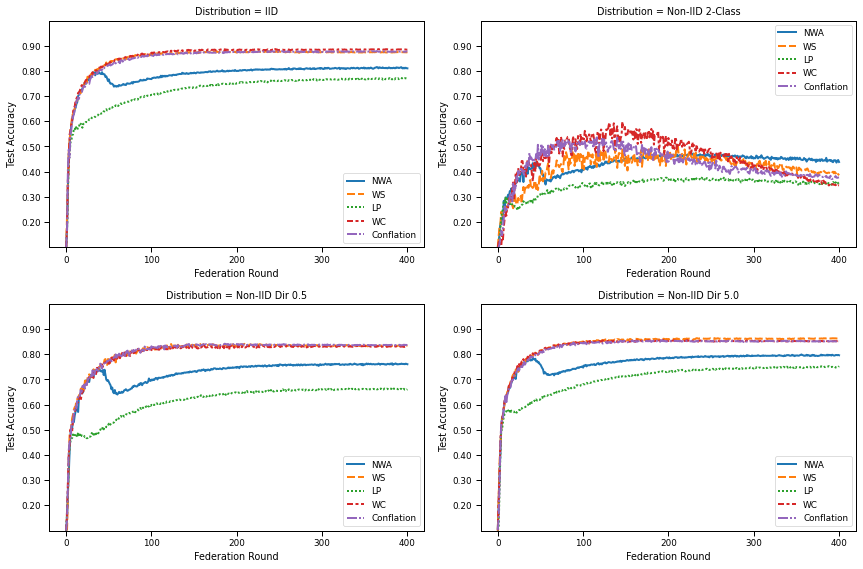}
    \caption{Learning curve plots for $E=1$. Plot of test set accuracy vs. federation round for \ac{VI} aggregation methods for each data distribution.}  
    \label{fig:E1_LearnCurveComparison}
\end{figure*}

In Fig. \ref{fig:E1_LearnCurveComparison}, we plot learning curve comparisons (test set accuracy vs federation round) for each of the data distributions.  Again, models trained using \ac{WS}, \ac{WC}, and conflation exhibit similar behavior.  The models trained using each of these aggregation methods converge much faster than the \ac{LP} and \ac{NWA} strategies.

\begin{table*}[ht!]
    \small
    \begin{center}
    \caption{Simulation Results - Site Epochs ($E$ = 1)}
    \resizebox{\textwidth}{!}{
        \label{tab:resultsLocalEpoch1}
        \begin{tabular}{|c | c |  c | c c c c c c c c|}
            \hline
        
            \multicolumn{3}{|c}{} & \multicolumn{2}{c}{\textbf{IID}} & \multicolumn{2}{c}{\textbf{Non-IID 2-Class}} & \multicolumn{2}{c}{\textbf{Non-IID Dir ($\alpha = 0.5$)}} & \multicolumn{2}{c|}{\textbf{Non-IID Dir ($\alpha = 5.0$)}} \\

            \hline
            \textbf{Model} & \textbf{Agg. Fn.} & \textbf{Client Weighting} & \textbf{Accuracy} & \textbf{NLL} & \textbf{Accuracy} & \textbf{NLL} & \textbf{Accuracy} & \textbf{NLL} & \textbf{Accuracy} & \textbf{NLL}\\
        
            \hline

            Deterministic & \ac{NWA} & Train Size& $87.66\pm0.001$ & $0.370\pm0.005$ & $51.45\pm0.020$ & $1.55\pm0.040$ & $83.45\pm0.003$ & $0.485\pm0.008$ & $85.78\pm0.003$ & $0.435\pm0.001$ \\
        
            \hline

            MC Dropout & \ac{NWA} & Train Size & $89.19\pm0.001$ & $0.330\pm0.006$ & $47.20\pm0.020$ & $1.65\pm0.020$ & $85.52\pm0.002$ & $0.431\pm0.004$ & $87.45\pm0.002$ & $0.373\pm0.005$ \\

            \hline

            \multirow{11}{*}{Flipout (\ac{VI})} & \multirow{3}{*}{\ac{NWA}} & Train Size & $81.67\pm0.002$ & $0.576\pm0.004$ & $43.04\pm0.025$ & $1.68\pm0.006$ & $74.77\pm0.008$ & $0.721\pm0.015$ & $79.89\pm0.002$ & $0.62\pm0.005$ \\
            & & Max Disc & $81.28\pm0.001$ & $0.581\pm0.002$ & $47.81\pm0.024$ & $1.65\pm0.013$ & $75.45\pm0.009$ & $0.722\pm0.005$ & $80.06\pm0.003$ & $0.608\pm0.012$ \\
            & & Distance & $81.76\pm0.003$ & $0.576\pm0.007$ & $40.00\pm0.030$ & $1.80\pm0.036$ & $75.92\pm0.011$ & $0.717\pm0.003$ & $80.04\pm0.001$ & $0.614\pm0.002$ \\

            \cline{2-11}
        
            & \multirow{3}{*}{\ac{WS}} & Train Size & $87.41\pm0.002$ & $0.388\pm0.006$ & $53.52\pm0.056$ & $1.41\pm0.076$ & $83.74\pm0.002$ & $0.508\pm0.014$ & $85.17\pm0.004$ & $0.465\pm0.01$ \\
            & & Max Disc & $87.81\pm0.004$ & $0.377\pm0.004$ & $47.72\pm0.049$ & $1.49\pm0.048$ & $84.17\pm0.004$ & $0.487\pm0.008$ & $84.90\pm0.002$ & $0.476\pm0.001$ \\
            & & Distance & $87.55\pm0.003$ & $0.391\pm0.004$ & $51.84\pm0.014$ & $1.47\pm0.021$ & $83.23\pm0.003$ & $0.51\pm0.005$ & $85.07\pm0.002$ & $0.464\pm0.005$ \\

            \cline{2-11}
        
            & \multirow{3}{*}{\ac{LP}} & Train Size & $77.21\pm0.004$ & $0.683\pm0.007$ & $35.50\pm0.048$ & $1.85\pm0.046$ & $67.67\pm0.008$ & $0.925\pm0.02$ & $74.7\pm0.007$ & $0.744\pm0.016$ \\
            & & Max Disc & $77.35\pm0.004$ & $0.681\pm0.012$ & $37.86\pm0.018$ & $1.79\pm0.014$ & $69.35\pm0.009$ & $0.895\pm0.017$ & $76.01\pm0.011$ & $0.717\pm0.023$ \\
            & & Distance & $77.69\pm0.002$ & $0.674\pm0.005$ & $32.48\pm0.064$ & $1.90\pm0.085$ & $68.47\pm0.015$ & $0.914\pm0.034$ & $75.94\pm0.005$ & $0.721\pm0.009$ \\

            \cline{2-11}
        
            & \multirow{3}{*}{\ac{WC}} & Train Size & $87.95\pm0.001$ & $0.377\pm0.006$ & $53.00\pm0.052$ & $1.44\pm0.096$ & $83.18\pm0.004$ & $0.510\pm0.009$ & $85.09\pm0.003$ & $0.478\pm0.001$ \\
            & & Max Disc & $87.81\pm0.002$ & $0.384\pm0.004$ & $50.71\pm0.025$ & $1.49\pm0.044$ & $83.70\pm0.002$ & $0.514\pm0.004$ & $84.86\pm0.002$ & $0.470\pm0.005$ \\
            & & Distance & $87.52\pm0.002$ & $0.389\pm0.002$ & $46.41\pm0.037$ & $1.52\pm0.037$ & $83.17\pm0.001$ & $0.514\pm0.006$ & $85.23\pm0.002$ & $0.464\pm0.006$ \\

            \cline{2-11}
        
            & Conflation & N/A & $87.42\pm0.001$ & $0.385\pm0.003$ & $53.09\pm0.008$ & $1.46\pm0.027$ & $83.18\pm0.001$ & $0.502\pm0.008$ & $84.94\pm0.005$ & $0.471\pm0.008$ \\

            \hline
        
        \end{tabular}}
  \end{center}
\end{table*}

\subsection{Evaluation of Aggregation Methods - 5 Local Epochs}
Next, we increase the amount of client computation and evaluate each of the proposed aggregation strategies with $E = 5$ local epochs.  Each simulation is run three times for 200 rounds, and the results, shown in Table~\ref{tab:resultsLocalEpoch5}, are averaged across experiments.

The results, depicted in Table~\ref{tab:resultsLocalEpoch5}, show much more parity between aggregation strategies in terms of accuracy and \ac{NLL}.  In fact, models trained using the \ac{NWA} strategy perform as well as, or better, than the other aggregation strategies across the board.  When compared to the $E=1$ experiment results (Table~\ref{tab:resultsLocalEpoch1}), models employing \ac{WS}, \ac{WC}, and conflation show minimal improvement when client computation is increased.  In some cases (\ac{WC}/conflation - non-IID 2-class), the models trained using $E=5$ actually perform worse.  Notably, none of the \ac{VI} aggregation methods perform as well as the deterministic baseline on the \ac{IID} or non-IID Dirichlet data distributions. However, all of them significantly improve on the deterministic baseline in the non-IID 2-class case.

\begin{figure*}[!ht]
    \centering
        \includegraphics[width=\textwidth]{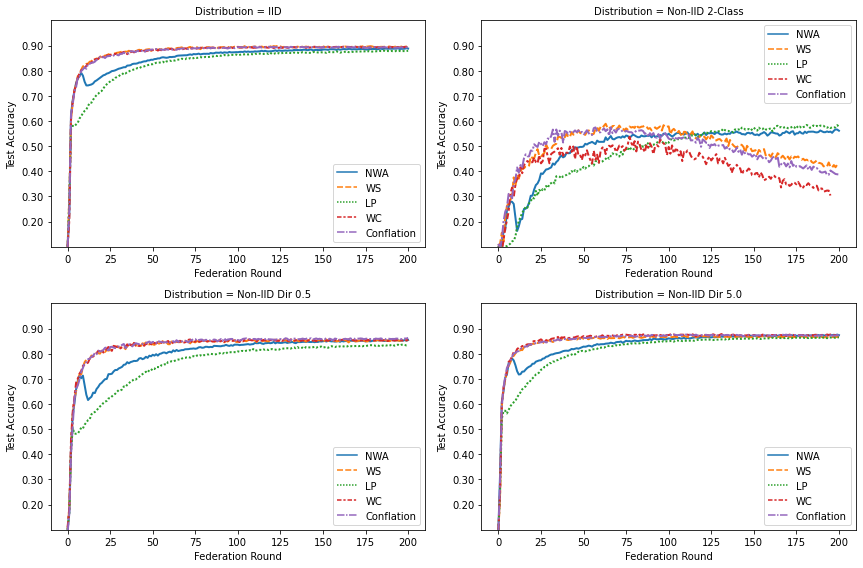}
    \caption{Learning curve plots for $E=5$. Plot of test set accuracy vs. federation round for \ac{VI} aggregation methods for each data distribution.}  
    \label{fig:E5_LearnCurveComparison}
\end{figure*}

Similar to the accuracy and \ac{NLL} results, the calibration differences across aggregation strategies are drastically reduced in the IID and non-IID Dirichlet cases. However, as is the case in the $E=1$ experiments, \ac{NWA} and \ac{LP} do not display a calibrated epistemic uncertainty component. The learning curve comparison plots in Fig.~\ref{fig:E5_LearnCurveComparison} display the same relationships as their $E=1$ counterparts in Fig.~\ref{fig:E1_LearnCurveComparison}.  

\begin{table*}[ht!]
    \small
    \begin{center}
    \caption{Simulation Results - Site Epochs ($E$ = 5)}
    \label{tab:resultsLocalEpoch5}

    \resizebox{\textwidth}{!}{

        \begin{tabular}{|c | c |  c | c c c c c c c c|}
            \hline
        
            \multicolumn{3}{|c}{} & \multicolumn{2}{c}{\textbf{IID}} & \multicolumn{2}{c}{\textbf{Non-IID 2-Class}} & \multicolumn{2}{c}{\textbf{Non-IID Dir ($\alpha = 0.5$)}} & \multicolumn{2}{c|}{\textbf{Non-IID Dir ($\alpha = 5.0$)}} \\

            \hline
            \textbf{Model} & \textbf{Agg. Fn.} & \textbf{Client Weighting} & \textbf{Accuracy} & \textbf{NLL} & \textbf{Accuracy} & \textbf{NLL} & \textbf{Accuracy} & \textbf{NLL} & \textbf{Accuracy} & \textbf{NLL}\\
        
            \hline

            Deterministic & \ac{NWA} & Train Size& $90.74\pm0.003$ & $0.371\pm0.009$ & $48.50\pm0.040$ & $1.56\pm0.060$ & $87.33\pm0.002$ & $0.456\pm0.008$ & $89.06\pm0.002$ & $0.456\pm0.008$ \\

            \hline

            MC Dropout & \ac{NWA} & Train Size & $91.16\pm0.002$ & $0.264\pm0.003$ & $44.90\pm0.040$ & $1.58\pm0.030$ & $87.91\pm0.001$ & $0.357\pm0.002$ & $89.53\pm0.002$ & $0.316\pm0.001$ \\

            \hline

            \multirow{11}{*}{Flipout (\ac{VI})} & \multirow{3}{*}{\ac{NWA}} & Train Size & $88.76\pm0.002$ & $0.334\pm0.005$ & $56.36\pm0.017$ & $1.55\pm0.021$ & $85.26\pm0.001$ & $0.439\pm0.001$ & $87.46\pm0.001$ & $0.380\pm0.004$ \\
            & & Max Disc & $88.87\pm0.001$ & $0.336\pm0.002$ & $55.91\pm0.061$ & $1.55\pm0.083$ & $85.43\pm0.001$ & $0.435\pm0.002$ & $87.30\pm0.003$ & $0.380\pm0.002$ \\
            & & Distance & $88.98\pm.003$ & $0.332\pm0.004$ & $45.60\pm0.129$ & $1.59\pm0.098$ & $85.35\pm0.003$ & $0.436\pm0.008$ & $87.38\pm0.001$ & $0.377\pm0.001$ \\

            \cline{2-11}
        
            & \multirow{3}{*}{\ac{WS}} & Train Size & $88.37\pm0.004$ & $0.446\pm0.008$ & $56.96\pm0.004$ & $1.36\pm0.015$ & $84.53\pm0.002$ & $0.585\pm0.004$ & $85.51\pm0.002$ & $0.545\pm0.002$ \\
            & & Max Disc & $88.23\pm0.005$ & $0.429\pm0.002$ & $49.84\pm0.036$ & $1.42\pm0.012$ & $84.68\pm0.004$ & $0.587\pm0.005$ & $84.92\pm0.009$ & $0.554\pm0.006$ \\
            & & Distance & $88.34\pm0.003$ & $0.444\pm0.003$ & $48.49\pm0.021$ & $1.42\pm0.017$ & $84.91\pm0.004$ & $0.581\pm0.005$ & $85.75\pm0.007$ & $0.545\pm0.014$ \\

            \cline{2-11}
        
            & \multirow{3}{*}{\ac{LP}} & Train Size & $88.05\pm0.002$ & $0.358\pm0.002$ & $55.41\pm0.023$ & $1.60\pm0.040$ & $83.59\pm0.001$ & $0.478\pm0.006$ & $86.36\pm0.001$ & $0.408\pm0.003$ \\
            & & Max Disc & $88.19\pm0.001$ & $0.356\pm0.001$ & $48.87\pm0.031$ & $1.65\pm0.041$ & $84.00\pm0.003$ & $0.472\pm0.008$ & $86.62\pm0.002$ & $0.404\pm0.004$ \\
            & & Distance & $87.91\pm0.001$ & $0.363\pm0.004$ & $41.95\pm0.050$ & $1.65\pm0.049$ & $84.03\pm0.001$ & $0.471\pm0.004$ & $86.47\pm0.001$ & $0.408\pm0.004$ \\

            \cline{2-11}
        
            & \multirow{3}{*}{\ac{WC}} & Train Size & $88.21\pm0.002$ & $0.443\pm0.002$ & $52.23\pm0.027$ & $1.41\pm0.040$ & $84.49\pm0.004$ & $0.583\pm0.006$ & $86.01\pm0.006$ & $0.542\pm0.014$ \\
            & & Max Disc & $88.34\pm0.002$ & $0.430\pm0.007$ & $49.21\pm0.051$ & $1.49\pm0.059$ & $83.52\pm0.002$ & $0.589\pm0.012$ & $84.47\pm0.014$ & $0.548\pm0.012$ \\
            & & Distance & $88.28\pm0.005$ & $0.437\pm0.012$ & $48.25\pm0.026$ & $1.44\pm0.043$ & $85.17\pm0.005$ & $0.574\pm0.009$ & $85.37\pm0.007$ & $0.578\pm0.006$ \\

            \cline{2-11}
        
            & Conflation & N/A & $88.16\pm0.008$ & $0.444\pm0.006$ & $52.47\pm0.028$ & $1.39\pm0.031$ & $84.86\pm0.003$ & $0.586\pm0.015$ & $85.59\pm0.006$ & $0.550\pm0.003$ \\              

            \hline
        
        \end{tabular}}
  \end{center}
\end{table*}

\begin{figure*}[!ht]
    \centering
        \subfloat{
            \centering
            \includegraphics[width=.48\textwidth]{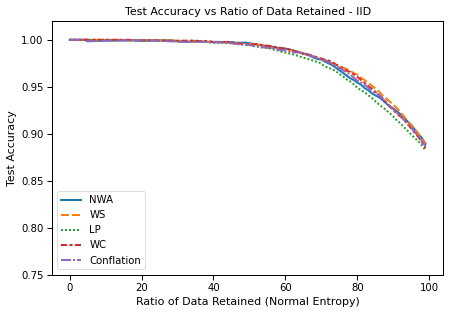}
        }
        \subfloat{
            \centering
            \includegraphics[width=.48\textwidth]{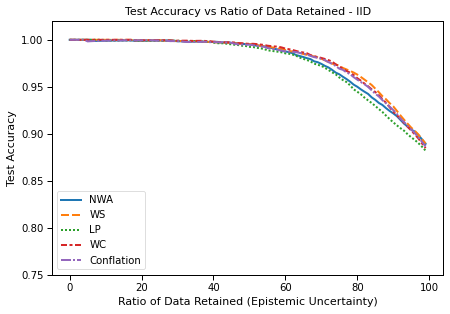}
        }\hfill%
        \subfloat{
            \centering
            \includegraphics[width=.48\textwidth]{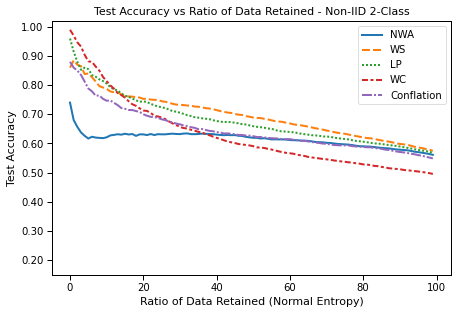}
        }
        \subfloat{
            \centering
            \includegraphics[width=.48\textwidth]{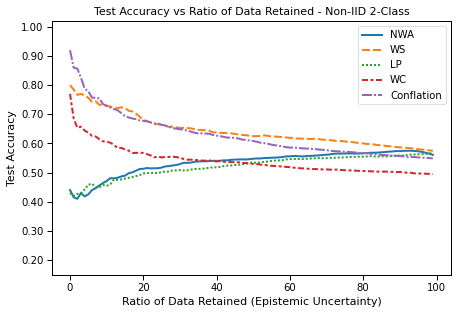}
        }\hfill%
    \caption{Uncertainty calibration plots for $E=5$.  \ac{VI} aggregation method comparison plot of test set accuracy vs. ratio of data retained for \ac{IID} and non-IID 2-class data distributions. Non-IID Dirichlet distribution plots show trends and relationships that mirror IID distribution results. Normalized entropy and epistemic uncertainty plots are shown, aleatoric uncertainty plots show trends and relationships that mirror normalized entropy.  }  
    \label{fig:E5_calibrationcomparison}
\end{figure*}

\subsection{Evaluation of Aggregation Methods - DWC and Model Pre-training}
The results of the \ac{DWC} experiments are presented separately due to significant differences in the training setup required by the \ac{DWC} aggregation strategy.  In Table~\ref{tab:resultsDWC}, we present \ac{DWC} results with and without global model pre-training.  For comparison, we employ the same pre-training strategy using the \ac{WS} aggregation method. Each of the DWC and model pre-training experiments is run three times for a single federation round, where each client trains for 400 local epochs.  Before distributing the global model to each client, the server pre-trains the global model on 10\% of the CIFAR-10 dataset for a maximum of 100 local epochs.  To avoid overfitting in our global model pre-training, we employ an early stopping strategy which monitors validation accuracy with a patience of 10 epochs.  

In Table~\ref{tab:resultsDWC}, the first row holds the results for \ac{DWC} without pre-training the global model.  In this case, \ac{DWC} does not hold up as a viable aggregation method.  One of the potential explanations for the poor performance of \ac{DWC} in this setup, is the use of a randomly initialized model as the initial global model. This is due to fact that the global model is heavily weighted in \ac{DWC}.  As a result, we experimented with pre-training the global model on a small subset of the overall training dataset.  This setup more closely resembles the experimental setup in~\citep{McClureDWC}.  Additionally, this is a likely scenario for a network of distributed sensors where a model is pre-trained on a centralized dataset and deployed on a group of sensors for collaborative learning.

\begin{figure*}[!ht]
    \centering
        \subfloat{
            \centering
            \includegraphics[width=.48\textwidth]{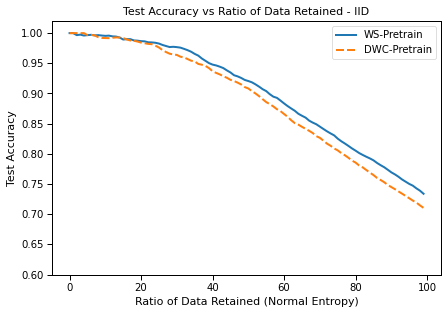}
        }
        \subfloat{
            \centering
            \includegraphics[width=.48\textwidth]{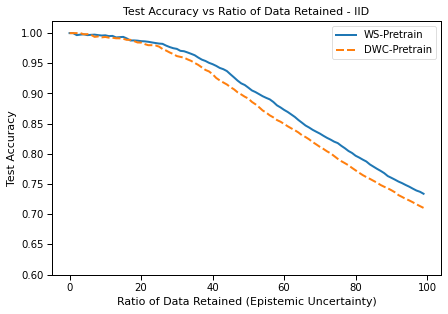}
        }\hfill%
        \subfloat{
            \centering
            \includegraphics[width=.48\textwidth]{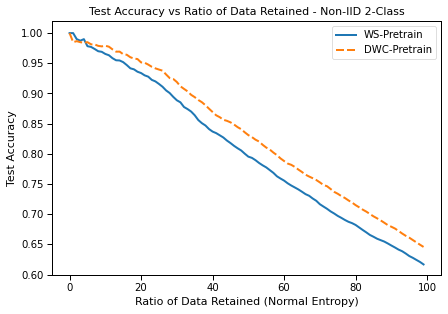}
        }
        \subfloat{
            \centering
            \includegraphics[width=.48\textwidth]{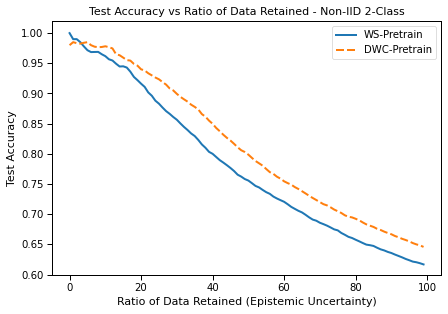}
        }\hfill%
    \caption{Uncertainty calibration plots for global model pre-training experiments.  \ac{DWC} and \ac{WS} aggregation method comparison plot of test set accuracy vs. ratio of data retained for \ac{IID} and non-IID 2-class data distributions. Non-IID Dirichlet distribution plots show trends and relationships that mirror IID distribution results. Normalized entropy and epistemic uncertainty plots are shown, aleatoric uncertainty plots show trends and relationships that mirror normalized entropy. }  
    \label{fig:DWCcalibrationplots}
\end{figure*}

In terms of accuracy and \ac{NLL}, our pre-training setup does not perform comparably to the $E=1$ or $E=5$ experiments in the \ac{IID} and Non-IID Dirichlet cases.  On the other hand, in the non-IID 2-class case, the models that were pre-trained perform exceedingly well.  In addition to a significant improvement in accuracy and \ac{NLL}, the uncertainty calibration plots in Fig.~\ref{fig:DWCcalibrationplots} display well-calibrated bulk (normalized entropy) and decomposed (aleatoric and epistemic) uncertainty components. 

\begin{table*}[ht!]
    \small
    \begin{center}
    \caption{Simulation Results - DWC}
    \label{tab:resultsDWC}

    \resizebox{\textwidth}{!}{
    
        \begin{tabular}{|c | c |  c | c c c c c c c c|}
            \hline
            
            \multicolumn{3}{|c}{} & \multicolumn{2}{c}{\textbf{IID}} & \multicolumn{2}{c}{\textbf{Non-IID 2-Class}} & \multicolumn{2}{c}{\textbf{Non-IID Dir ($\alpha = 0.5$)}} & \multicolumn{2}{c|}{\textbf{Non-IID Dir ($\alpha = 5.0$)}} \\

            \hline
            \textbf{Model} & \textbf{Agg. Fn.} & \textbf{Pre-train} & \textbf{Accuracy} & \textbf{NLL} & \textbf{Accuracy} & \textbf{NLL} & \textbf{Accuracy} & \textbf{NLL} & \textbf{Accuracy} & \textbf{NLL}\\
        
            \hline

            Flipout & \ac{DWC} & No & $34.95\pm 1.230$ & $1.73\pm 0.031$ & $11.64\pm 1.73$ & $2.69\pm0.220$ & $17.25\pm 3.14$ & $2.440\pm0.210$ & $31.83\pm1.05$ & $1.801\pm0.029$ \\

            \hline
        
            Flipout & \ac{DWC} & Yes & $71.42\pm 0.270$ & $0.83\pm 0.007$ & $66.35\pm 1.53$ & $0.97\pm0.038$ & $68.81\pm 0.42$ & $0.918\pm0.007$ & $65.87\pm1.36$ & $0.972\pm0.026$  \\

            \hline

            Flipout & \ac{WS} & Yes & $72.70\pm 0.008$ & $0.80\pm 0.017$ & $64.63\pm 1.02$ & $1.01\pm0.051$ & $62.65\pm0.02$ & $1.061\pm0.061$ & $68.18\pm0.02$ & $0.913\pm0.051$  \\
         
            \hline
        
        \end{tabular}}
  \end{center}
\end{table*}

\subsection{Updating the Prior Every Federation Round}
In this experiment, we assess the effect of updating the prior after every federation round.  Each of these experiments are run twice on the \ac{IID} data partition, and the results are averaged across experiments. The results in Table~\ref{tab:resultspriorupdate} show a precipitous drop in accuracy and \ac{NLL} when compared to the results in Tables~\ref{tab:resultsLocalEpoch1} and~\ref{tab:resultsLocalEpoch5}, demonstrating that updating the prior every federation round is not an effective strategy.

\begin{table*}[ht!]
    \small
    \begin{center}
    \caption{Simulation Results - Prior Updated Every Federation Round ($E = 1$)}
    \label{tab:resultspriorupdate}
    \resizebox{.5\textwidth}{!}{
        \begin{tabular}{|c | c |  c | c c |}
            \hline
        
            \multicolumn{3}{|c}{} & \multicolumn{2}{c|}{IID}   \\

            \hline
        
            \textbf{Model} & \textbf{Agg. Fn.} & \textbf{Client Weighting} & \textbf{Accuracy} & \textbf{NLL} \\
        
            \hline

            \multirow{5}{*}{Flipout (\ac{VI})} & \ac{NWA} & Train Size & $41.84\pm.59$ & $1.55\pm.009$  \\        

            \cline{2-5}
        
            & \ac{WS} & Train Size & $43.54\pm.54$ & $1.53\pm.017$ \\

            \cline{2-5}
        
            & \ac{LP} & Train Size & $46.14\pm.50$ & $1.46\pm.011$  \\

            \cline{2-5}
        
            & \ac{WC} & Train Size & $44.23\pm.52$ & $1.51\pm.002$ \\

            \cline{2-5}
        
            & Conflation & N/A & $42.62\pm.58$ & $1.56\pm.007$   \\

            \hline
        
        \end{tabular}}
  \end{center}
\end{table*}

\subsection{Client Weighting Strategies}
Results of the $E=1$ and $E=5$ experiments with different client weighting strategies are depicted in Tables~\ref{tab:resultsLocalEpoch1} and~\ref{tab:resultsLocalEpoch5}.  These experiments were conducted in conjunction with the $E=1$ and $E=5$ experiments, and utilized the same setup. Unfortunately, this experiment was inconclusive, and the results did not elucidate the importance of the different client weighting techniques.

\subsection{MC Dropout as a Lightweight Alternative to Mean-field Gaussian VI}
The \ac{MC} dropout experiments were conducted in conjunction with the $E=1$ and $E=5$ experiments and the setup is the same.  Accuracy and \ac{NLL} results are presented in Tables~\ref{tab:resultsLocalEpoch1} and~\ref{tab:resultsLocalEpoch5}.  

For both $E=1$ and $E=5$, in terms of accuracy/\ac{NLL}, the \ac{MC} dropout models using \ac{NWA} perform similarly to the deterministic baseline.  In the \ac{IID} and non-IID Dirichlet cases, the \ac{MC} dropout models outperform all the \ac{VI} aggregation methods.  On the other hand, in the non-IID 2-class case, the \ac{MC} dropout model does not perform nearly as well as the top-performing \ac{VI} aggregation methods.

\begin{figure*}[!ht]
    \centering
        \subfloat{
            \centering
            \includegraphics[width=.48\textwidth]{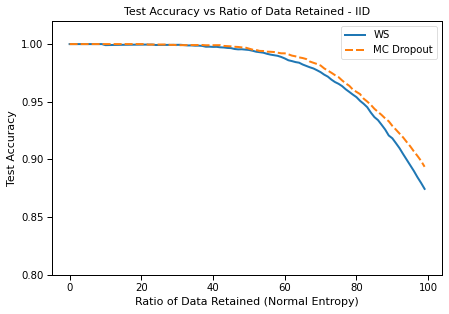}
        }
        \subfloat{
            \centering
            \includegraphics[width=.48\textwidth]{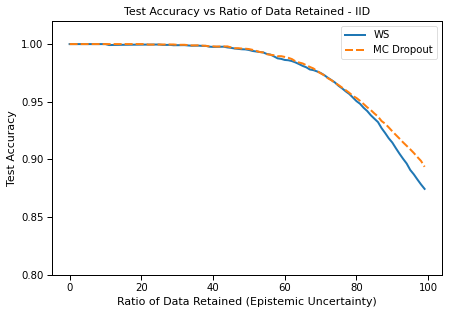}
        }\hfill%
        \subfloat{
            \centering
            \includegraphics[width=.48\textwidth]{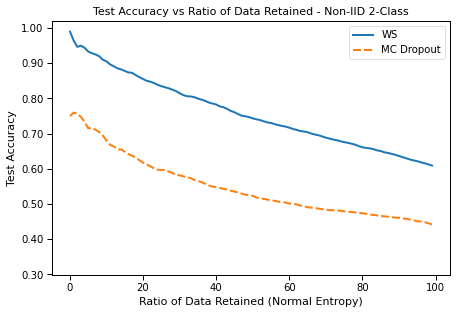}
        }
        \subfloat{
            \centering
            \includegraphics[width=.48\textwidth]{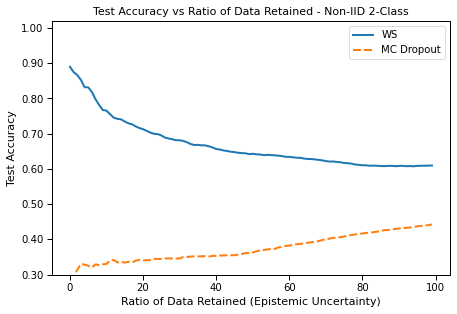}
        }\hfill%
    \caption{Uncertainty calibration plots for \ac{MC} dropout experiments ($E=1$).  \ac{MC} dropout (\ac{NWA}) and \ac{VI} (\ac{WS}) aggregation method comparison plot of test set accuracy vs. ratio of data retained for \ac{IID} and non-IID 2-class data distributions. Non-IID Dirichlet distribution plots show trends and relationships that mirror IID distribution results. Normalized entropy and epistemic uncertainty plots are shown, aleatoric uncertainty plots show trends and relationships that mirror normalized entropy.}  
    \label{fig:dropoutcalibrationplot}
\end{figure*}

In Fig.~\ref{fig:dropoutcalibrationplot}, we present uncertainty calibration comparison plots for $E=1$.  Like the previous uncertainty calibration plots, the non-IID Dirichlet plots depict the same trends and relationships as the \ac{IID} plots and are not shown.  The $E=5$ calibration plots also follow similar relationships and trends to the $E=1$ plots and are not shown.  In both cases, the models trained on \ac{IID} and non-IID Dirichlet data distributions appear to be well-calibrated.  In contrast, in the non-IID 2-class case, the \ac{MC} dropout models show poor calibration of the epistemic uncertainty.

\begin{figure*}[!ht]
\centering
    \includegraphics[width=\textwidth]{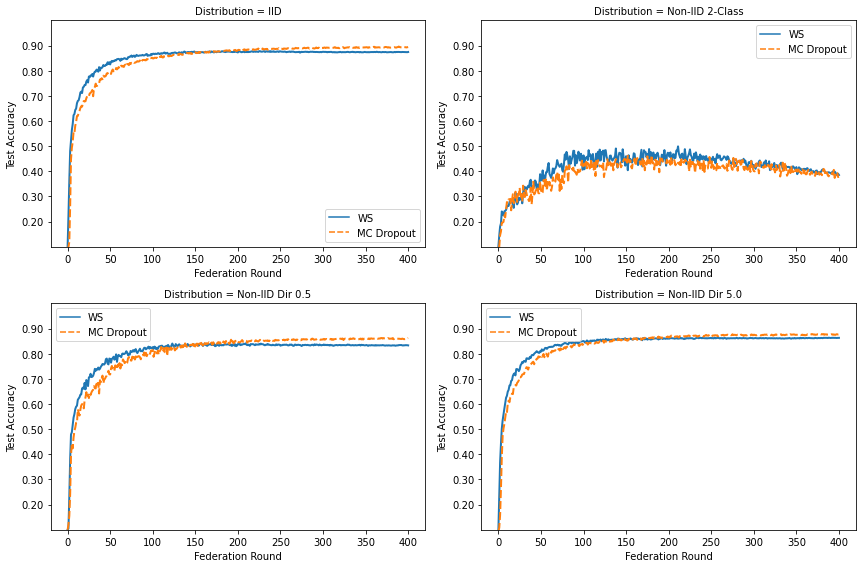}
    \caption{Learning curve plots for \ac{MC} dropout experiments ($E=1$). Plot of test set accuracy vs. federation round for \ac{MC} dropout (\ac{NWA}) and \ac{VI} (\ac{WS}) aggregation methods for each data distribution.}  
    \label{fig:dropout_LearnCurveComparison}
\end{figure*}

\begin{figure*}[!ht]
    \centering
        \includegraphics[width=\textwidth]{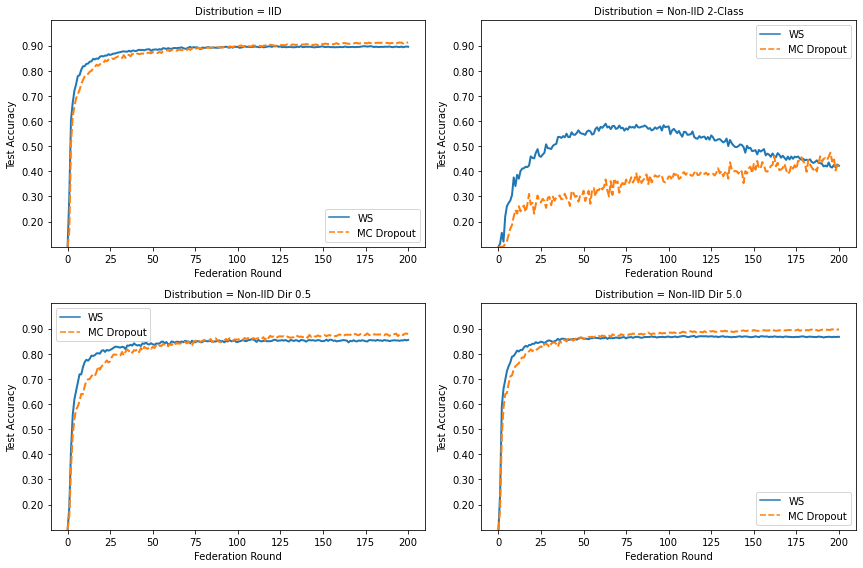}
    \caption{Learning curve plots for \ac{MC} dropout experiments ($E=5$). Plot of test set accuracy vs. federation round for \ac{MC} dropout (\ac{NWA}) and \ac{VI} (\ac{WS}) aggregation methods for each data distribution.}  
    \label{fig:dropout_LearnCurveComparison_E5}
\end{figure*}

Figures~\ref{fig:dropout_LearnCurveComparison} and~\ref{fig:dropout_LearnCurveComparison_E5} depict the learning curve comparisons for $E=1$ and $E=5$.  Similar to the calibration plots, \ac{WS} is used as a reference for comparison.  For $E=1$, the \ac{MC} dropout models converge at a marginally slower rate.  In the non-IID 2-class case for $E=5$, shown in Fig.~\ref{fig:dropout_LearnCurveComparison_E5}, the \ac{MC} dropout model converges at a significantly slower rate than the mean-field Gaussian \ac{VI} model trained using \ac{WS}.

\section{Discussion}
  In our evaluation of six different aggregation methods, and three client weighting schemes, we demonstrated that the results vary based on the training data distribution and number of local training epochs per federation round.  We showed that the selection of the aggregation method is a critical hyperparameter, with downstream effects on model performance (accuracy), calibration, and training time, which must be selected based on the machine learning model type, dataset, and other hyperparameters.  Notably, we established that often these aggregation strategies can be placed into two groups, with each group exhibiting similar performance (accuracy/\ac{NLL}), calibration, and convergence rate.  The first group consists of \ac{WS}, \ac{WC}, and conflation.  The second group consists of \ac{NWA} and \ac{LP}.  Models employing the aggregation strategies in the first group tend to exhibit relatively high accuracy, low \ac{NLL}, and rapid convergence with a single local epoch of training during each federation round.  These models are also better calibrated than their counterparts from the second group, which require five local epochs of training per federation round to reach the same performance level.  These relationships may facilitate a targeted hyperparameter search.  Specifically, the similarities amongst strategies in each group may allow the testing of a single strategy from each group to narrow the search.  

Although our experiments on client weighting techniques were not fruitful, we believe that this may be due to the CIFAR-10 dataset and the model we selected.  In \ac{FL} implementations with more complex architectures and complex, real-world, datasets, where distributions are highly non-IID, these weighting techniques may help to smooth training and reduce the effects of a non-representative client or an adversary attempting to disrupt the training process.  Further experimentation with larger and more complex models and datasets is necessary to validate these beliefs.  

The \ac{DWC} algorithm required some significant adjustments to our experimental setup; however, it drove us to experiment with pre-training.  In many remote sensing tasks, like visual target recognition or passive sonar target classification~\citep{FischerJoE}, this setup is likely.  Prior to deployment on a remote sensing platform, the model needs to be trained and evaluated on a representative dataset until it demonstrates an acceptable level of performance.  The random initialization is not realistic in these remote sensing applications.  Once the model is deployed on a group of sensors, it can be fine-tuned using \ac{FL} in a setup that mirrors our experimental setup.  In this application, \ac{DWC} is a viable solution and displays comparable performance to \ac{WS}.  

The cost of deploying \ac{VI} models to remote sensing platforms and training those models, in many cases, is prohibitive. \ac{MC} dropout is a lightweight alternative to mean-field Gaussian \ac{VI} which, prior to this work, has not been directly compared to \ac{VI} in the \ac{FL} environment.  In our experiments, we showed that in the \ac{IID} and non-IID Dirichlet cases, \ac{MC} dropout performs on par with one of the top performing \ac{VI} aggregation methods (\ac{WS}).  In fact, in terms of accuracy and \ac{NLL}, the \ac{MC} dropout models outperform the mean-field Gaussian \ac{VI} models.  Although the mean-field Gaussian \ac{VI} models tended to converge faster, the \ac{MC} dropout models have half the number of parameters, which require fewer client resources and lower communication requirements between client and server.  The non-IID 2-class data distribution case was the exception: the \ac{MC} dropout models converged slower, showed a reduction in accuracy, and displayed poor calibration.  In some cases, the use of an \ac{MC} dropout model utilizing the \ac{NWA} aggregation strategy may be an acceptable lightweight alternative to mean-field Gaussian \ac{VI} models employing one of the aggregation strategies discussed in this manuscript.  

\section{Conclusion and Future Work}
In remote sensing and safety-critical applications, the ability of a model to communicate a measure of epistemic uncertainty is essential.  \ac{BDL} models are capable of communicating a measure of epistemic uncertainty while simultaneously delivering competitive prediction accuracy when compared to their deterministic counterparts.  However, to fully realize the benefits of \ac{BDL} in the \ac{FL} setting, it is important to understand the downstream effects of the selected model aggregation method.  In this manuscript, we used four different partitions of the CIFAR-10 dataset and a fully variational ResNet-20 architecture to analyze six \ac{BFL} model aggregation methods for mean-field Gaussian \ac{VI} models. This is the first time, to our knowledge, that a fully variational architecture (i.e., all model layers are variational) has been utilized in this type of analysis. Additionally, when trained on a centralized dataset, each of the variants of our ResNet-20 architecture achieved comparable performance to the benchmark performance reported by He \textit{et al.}~\citep{he2016deep}, see Table~\ref{tab:centralizedresults}. As such, we were able to compare our \ac{FL} results to an important \ac{ML} benchmark and provide a more detailed look at uncertainty.  Ultimately, we identified the aggregation method as a critical hyperparameter in the development of a \ac{BFL} strategy, and we showed that, in certain applications, \ac{MC} dropout is an acceptable lightweight alternative to mean-field Gaussian \ac{VI}.  In both cases (mean-field Gaussian \ac{VI} and \ac{MC} dropout), our global models (Table~\ref{tab:resultsLocalEpoch5}, $E=5$) achieved test set accuracy within $\pm{2-3\%}$ of the centralized benchmark.

\section{Disclaimer}
The views and conclusions contained herein are those of the authors and should not be interpreted as necessarily representing the official policies or endorsements, either expressed or implied, of the U.S. Government. The U.S. Government is authorized to reproduce and distribute reprints for Government purposes notwithstanding any copyright annotations thereon.

\bibliographystyle{IEEEbib}
\bibliography{refs}
\end{document}